\documentclass[10pt,twocolumn,letterpaper]{article}

\usepackage[pagenumbers]{cvpr} 
\usepackage{graphicx}
\usepackage{amsmath}
\usepackage{amssymb}
\usepackage{booktabs}
\usepackage{tabularx}
\usepackage{algorithm}
\usepackage{algorithmic}
\usepackage[dvipsnames]{xcolor}

\definecolor{cvprblue}{rgb}{0.21,0.49,0.74}
\usepackage[pagebackref,breaklinks,colorlinks,citecolor=cvprblue]{hyperref}

\usepackage[capitalize]{cleveref}
\crefname{section}{Sec.}{Secs.}
\Crefname{section}{Section}{Sections}
\Crefname{table}{Table}{Tables}
\crefname{table}{Tab.}{Tabs.}

\def\paperID{7181} 
\def\confName{CVPR}
\def\confYear{2024}

\title{Enhancing Prompt Following with Visual Control\\ Through Training-Free Mask-Guided Diffusion}

\author{
    Hongyu Chen$^{1}$\thanks{Both authors contributed equally to this work.}\quad
    Yiqi Gao$^{2}$\textsuperscript{*}\quad
    Min Zhou$^{1}$ \quad
    Peng Wang$^{2}$\quad
    Xubin Li$^{1}$\quad \\
    Tiezheng Ge$^{1}$\quad
    Bo Zheng $^{1}$ \\
    \{yinchen.chy, yunqi.zm, lxb204722, tiezheng.gtz, bozheng\}@alibaba-inc.com \\
    \{gyqjz, peng.wang\}@nwpu.edu.cn \\
    $^1$Alibaba Group\quad \\
    $^2$Northwestern Polytechnical University
}

\begin{document}



\maketitle

\begin{abstract}


Recently, integrating visual controls into text-to-image~(T2I) models, such as ControlNet method, has received significant attention for finer control capabilities. While various training-free methods make efforts to enhance prompt following in T2I models, the issue under visual control is still rarely studied, especially in the scenario that visual controls are misaligned with text prompts. In this paper, we address the challenge of ``Prompt Following With Visual Control" and propose a training-free approach named Mask-guided Prompt Following (MGPF). Object masks are introduced to distinct aligned and misaligned parts of visual controls and prompts. Meanwhile, a network, dubbed as Masked ControlNet, is designed to utilize these object masks for object generation in the misaligned visual control region. Further, to improve attribute matching, a simple yet efficient loss is designed to align the attention maps of attributes with object regions constrained by ControlNet and object masks. The efficacy and superiority of MGPF are validated through comprehensive quantitative and qualitative experiments.

\end{abstract}

\section{Introduction}

\label{sec:intro}

Adding image conditions to text-to-image~(T2I) models~\cite{mou2023t2i, huang2023composer, zhang2023adding} for accurate layout and shape control has attracted significant attention. Notably, ControlNet~\cite{zhang2023adding} stands out as a widely adopted solution for its effectiveness in producing high-quality images with diverse visual controls such as Canny Edge~\cite{canny1986computational}, Depth Map~\cite{ranftl2020towards}, and other form images. Though many training-free methods~\cite{chefer2023attend, hertz2022prompt, bar2022text2live, avrahami2023blended, nichol2021glide,xie2023boxdiff, agarwal2023star, Cao_Wang_Qi_Shan_Qie_Zheng_2023, Liu_Li_Du_Torralba_Tenenbaum_2022, Li_Keuper_Zhang_Khoreva_2023, Park_Kim_Kim_Lee_Ye_2023} have been proposed to make generated images consistent with text prompts in T2I models, addressing the ``Prompt Following" issue~\cite{betker2023dalle3}, research on this issue within ControlNet is rare. This paper delves into this unexplored area, labeled as ``Prompt Following Under Visual Control".

\begin{figure}[ht!]
		\begin{center}
			\includegraphics[width=1.\linewidth]{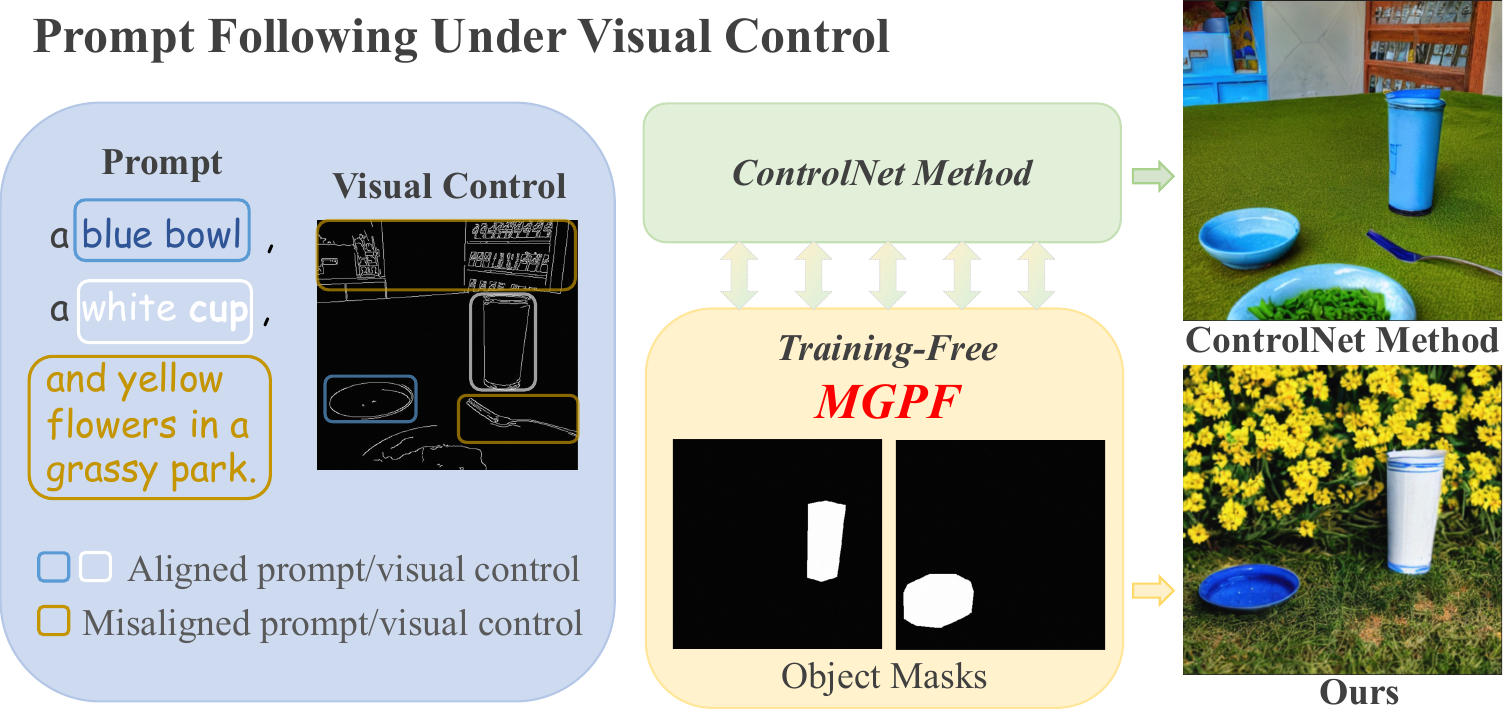}
		\end{center}
		\caption{The input prompt and visual control are only partially aligned as the left part displays. The ControlNet method result omits yellow flowers and a grassy park, inaccurately painting the cup as blue not white. While our approach MGPF can utilize object masks and produce images with these desired features.}
		\label{showcase1}
\end{figure}


 The key challenge lies in ensuring the generated images match not only the text prompts but also the layouts and shapes specified by the visual controls. This task becomes more crucial when users can only provide partially aligned prompt-image pairs since it is hard to find visual controls exactly matching their prompts.
 For example, as~\cref{showcase1} shows, the canny edge condition derived from an indoor setting only partially aligns with the prompt. This misalignment influences the effectiveness of the prompt and results in missing elements like flowers and grass in the generated image~(termed as ``Object Missing"). Moreover, the attributes of objects are incorrect, such as the color of the cup being blue instead of white~(termed as ``Attribute Mismatch").  This issue may arise from the inconsistency between the attention maps of the attribute words and that of the corresponding object during the denoising process~\cite{tang2022daam}. 

Directly transferring prompt following methods from T2I to the ControlNet scenario does not yield satisfactory results. Specifically, in optimizing Object Missing, some works \cite{chefer2023attend, hertz2022prompt, bar2022text2live, avrahami2023blended, nichol2021glide,xie2023boxdiff} focus on augmenting the attention values of object words in the prompt during cross-attention, proving effective in T2I scenarios. However, when applied to ControlNet~\cite{zhang2023adding}, these methods still suffer from the misalignment of prompts and visual controls, leading to a weakened effect. In addressing Attribute Mismatch, some studies\cite{rassin2023linguistic, chefer2023attend, Li_Keuper_Zhang_Khoreva_2023} suggest bringing the attention maps of attribute words and object words within U-net closer to each other.
This process may affect the original layout distribution of objects. Such approaches are effective in T2I scenarios without layout and shape constraints, but may conflict with the visual controls in ControlNet.

To address these issues, we introduce additional masks for objects aligned with prompts, and propose a novel training-free approach named Mask-guided Prompt Following (\textbf{MGPF}). Regarding the ``Object Missing" challenge, we introduce Masked ControlNet to replace the original ControlNet branch and mitigate conflicts between prompts and the misaligned part of visual controls. 
Specifically, we employ object masks to divide ControlNet features into two parts, only integrating features aligned with the prompt into the U-net. 
We note that partial ControlNet features can effectively control particular spatial compositions in the generated image without disrupting non-adjacent regions or introducing artifacts. This is likely because the U-net remains fixed during training, preserving its ability to generate images. Also, combining ControlNet features with U-net features maintains spatial information in the ControlNet features. Additionally, ControlNet features are combined with U-net features through element-wise addition, preserving the spatial information from visual controls. 

\begin{figure}
		\begin{center}
			\includegraphics[width=1.\linewidth]{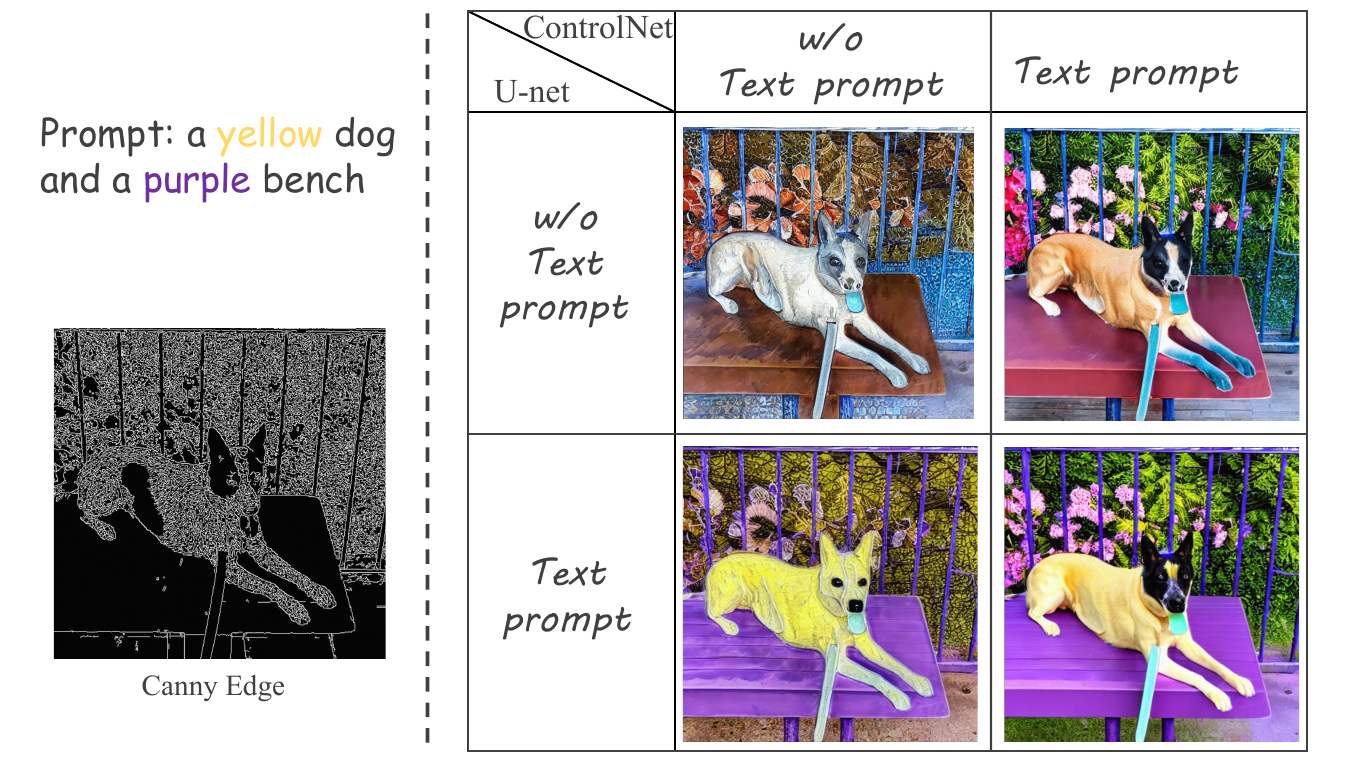}
		\end{center}
		\caption{Under visual control, the prompt is respectively fed into or unfed into the U-net and ControlNet. If U-net does not received the prompt, regardless of whether ControlNet has received it or not, the image lacks yellow and purple as mentioned. But if U-net gets the prompt, it is the opposite. This reveals that attribute words mostly work through the cross-attention between U-net and the prompt features.}
		\label{insight}
\end{figure}

Furthermore, we propose Attribute Matching loss for binding attributes to objects and avoid disturbing visual controls. As depicted in~\cref{insight}, we give different prompts to U-net and ControlNet and find that the cross-attention between U-net and text features determines the attribute of objects while ControlNet predominantly determines the layout and shape of objects. Inspired by this, we utilize classifier guidance~\cite{DBLP:conf/nips/DhariwalN21} and minimize the divergence distance between attention maps of attribute words in Stable Diffusion and those of object words in ControlNet. Additionally, we impose a constraint to ensure that the regions of highest attention for attribute words and corresponding object words within the object mask regions. For the misaligned prompt and visual control, we just make the attention maps for attributes and objects within the U-net consistent.

To evaluate our method, we extract various visual control and object masks from the COCO dataset~\cite{lin2014microsoft} and design paired prompts that include a variety of attributes and objects. The experiments demonstrate that our model outperforms state-of-the-art methods. 

In summary, our works can be summarized as: 
(i) We explore the challenge of prompt following under visual control and introduce object masks for better semantic expression.
(ii) We propose a training-free method called Masked-guided Prompt Following, including Masked ControlNet and Attribute-matching Loss for precise object and attribute control. 
(iii) Extensive comparisons show that our method is simple yet effective for controllable text-to-image generation within multiple conditions.

\section{Related Works}
\noindent\textbf{Text-to-Image Generation.} 
Early research in text-to-image generation primarily utilized Generative Adversarial Networks (GANs), as demonstrated in studies by Tao Tang et al.~\cite{Tao_Tang_Wu_Sebe_Jing_Wu_Bao}, Xu et al.~\cite{Xu_Zhang_Huang_Zhang_Gan_Huang_He_2018}, and Zhang et al.~\cite{Zhang_Koh_Baldridge_Lee_Yang_2021}. However, recent advancements have shifted focus towards diffusion models~\cite{Ramesh_Dhariwal_Nichol_Chu_Chen, Ramesh_Pavlov_Goh_Gray_Voss_Radford_Chen_Sutskever_2021, Esser_Rombach_Ommer_2021, Balaji_Nah_Huang_Vahdat_Song_Kreis_Aittala_Aila_Laine_Catanzaro_et_al._2022, ho2020denoising, song2020denoising, nichol2021improved}, gaining prominence for their exceptional capabilities in generating high-quality and high-resolution images. A notable example in this domain is the Denoising Diffusion Probabilistic Model (DDPM)~\cite{ho2020denoising}, which introduces standard noise in its forward process and reconstructs the original image from the noisy data in the reverse process. In this reverse process, a U-Net-based model~\cite{Ronneberger_Fischer_Brox_2015} is employed to estimate the noise at each step. This era is further marked by groundbreaking large-scale text-to-image architectures such as Stable Diffusion~\cite{rombach2022high}, Imagen by Saharia et al.~\cite{saharia2022photorealistic}, and DALL·E 3 by OpenAI~\cite{betker2023dalle3}. Besides, these models have been applied to a diverse range of tasks beyond those initially mentioned, including 3D generation~\cite{Poole_Jain_Barron_Mildenhall_Research_Berkeley, Fridman_Abecasis_Kasten_Dekel_2023, ollein_Cao_Owens_Johnson_Niener_2023}, video editing~\cite{Geyer_Bar-Tal_Bagon_Dekel_2023, Liu_Zhang_Li_Lin_Jia_2023, Molad_Horwitz_Valevski_Acha_Matias_Pritch_Leviathan_Hoshen_2023, Qi_Cun_Zhang_Lei_Wang_Shan_Chen_2023, Yang_Zhou_Liu_Loy} and motion generation~\cite{Raab_Leibovitch_Tevet_Arar_Bermano_Cohen-Or_2023, Tevet_Raab_Gordon_Shafir_Bermano_Cohen-Or_2022}. Building upon these state-of-the-art models, our work aims to enhance the prompt-following capabilities of these systems under controlled conditions.

\noindent\textbf{Prompt Following.} Generating images that precisely align with complex text prompts remains a formidable challenge, as proved in recent studies~\cite{chefer2023attend, feng2022training, hertz2022prompt, Wang_Chen_Chen_Ma_Lu_Lin}. This issue is particularly pronounced in scenarios involving complex scenes or multi-object compositions~\cite{Ma_Lewis_Kleijn_Leung_2023, feng2022training, chefer2023attend}. StructureDiffusion~\cite{feng2022training} attempts to address this by leveraging linguistic structures to guide cross-attention maps in the generation process but falls short in resolving semantic discrepancies at the sample level. Composable Diffusion~\cite{Liu_Li_Du_Torralba_Tenenbaum_2022} decomposes complex texts into simpler segments and then composes the image from these segments, though its effectiveness is limited to simpler conjunctions. Additionally, A\&E~\cite{chefer2023attend} aims to enhance object presence in generated images by optimizing cross-attention maps during inference. However, it struggles with intricate prompts, particularly those with complex backgrounds, and does not address attribute-related issues effectively. In response, Linguistic Diffusion~\cite{rassin2023linguistic} introduces a specialized KL-divergence loss to fine-tune the cross-attention maps of modifiers and entities, making significant progress. Despite these advancements, existing methods primarily focus on general text-to-image model improvements. In contrast, our paper targets enhancing prompt following under visual control, an area where previous approaches have shown limitations.

\noindent\textbf{Controlling Text-to-Image Diffusion Models.} The state-of-the-art diffusion models primarily discussed earlier are geared towards text-to-image generation, relying heavily on text prompts for control. However, achieving precise control solely through text prompts presents inherent challenges. Recent developments have introduced methods to exert additional visual control over existing text-to-image models. Innovations like ControlNet~\cite{zhang2023adding} and T2I Adapter~\cite{mou2023t2i} facilitate this enhanced control, enabling models such as Stable Diffusion~\cite{rombach2022high}~(SD) to adapt to various image contexts with minimal additional training. ControlNet~\cite{zhang2023adding} integrates supplementary features from diverse image settings into the core SD U-Net model, while T2I Adapters~\cite{mou2023t2i} employ a different approach in incorporating these features and managing inputs. Our research focuses on scenarios using ControlNet~\cite{zhang2023adding}, where we observe a notable decline in the quality of prompt following compared to general scenarios. To the best of our knowledge, ours is the first work addressing this specific challenge. We propose a novel training-free method to resolve this issue without compromising flexibility.

\section{Method}

\begin{figure*}
		\begin{center}
			\includegraphics[width=0.95\linewidth]{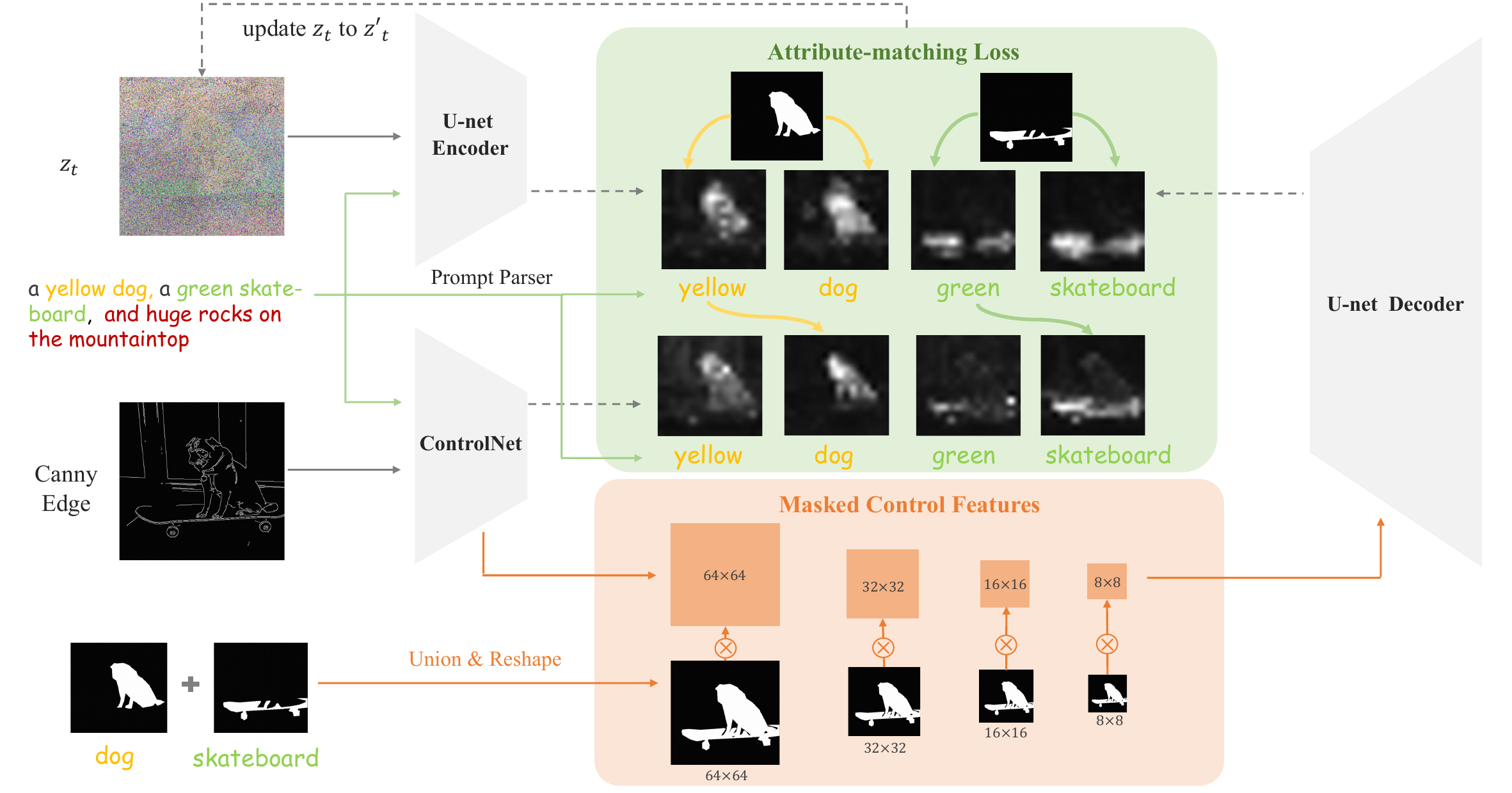}
		\end{center}
		\caption{Overview of the proposed Mask-guided Prompt Following (MGPF) method. Given a prompt and a canny edge condition with misaligned elements such as door edges, along with two object masks indicating ``dog" and ``skateboard", our approach involves two modules to enhance prompt following. In Masked ControlNet, We union all object masks into a single composite, being reshaped and element-wise multiplied to corresponding ControlNet features, effectively eliminating the influence of undesired visual clues. Incorporated with Attribute-matching Loss, we parse the prompt into attribute-object pairs like ``yellow dog" and ``green skateboard", obtaining their cross-attention maps from U-net and ControlNet. Subsequently, specific loss functions shift these attention maps in U-net and ControlNet for better attribute binding.}
		\label{overview}
\end{figure*}

This section elaborates on the proposed Mask-guided Prompt Following (\textbf{MGPF}). As illustrated in~\cref{overview}, this method adopts SD for achieving high-quality text-to-image generation and ControlNet to integrate diverse visual controls. MGPF consists of two key components: 1. Masked ControlNet, which leverages object masks to control the layout and shape of specific regions, thereby eliminating conflicts with misaligned portions of the prompt and enhancing object generation; 2. Attribute-matching Loss, which employs related loss functions to bind attributes within the prompt to their corresponding objects. The text prompt$\mathcal{P}=\{p\}_{i=1}^L$and image condition $I$ serve as fundamental inputs for controlled image synthesis, where $L$ denotes the length of prompt. 
 In addition, object masks $\mathcal{M}=\{m\}_{i=1}^N$  are provided, where $m_{i}$ denotes pixel indices associated with the $i_{th}$ object described in $\mathcal{P}$ and $I$.



\subsection{Masked ControlNet}

To address the challenge of generating objects in areas where the prompt and visual controls are misaligned,  
We propose the Masked ControlNet, which is then integrated with Stable Diffusion(SD)~\cite{rombach2022high}.
SD, built on a U-net architecture, includes an encoder, a middle block and a decoder, all interconnected via skip connections. It utilizes the prompt as a basis condition for text-to-image generation. To incorporate more visual controls, ControlNet~\cite{zhang2023adding} is developed using a trainable copy of the 12 encoding blocks and 1 middle block of SD.  The outputs from ControlNet are subsequently merged into the 12 skip connections and the middle block of the U-net, thereby serving as supplementary conditions for the generation process.
Consequently, when presented with a textual prompt $\mathcal{P}$ and an image condition $I$, the computation of the score for classifier-free guidance~\cite{ho2022classifier} in multi-conditions proceeds as follows:
\begin{equation}
		\begin{split}
			\label{sd_control}
                &\epsilon_{CFG}(z_{t}, \mathcal{P}, I)=\epsilon_{\theta}(z_{t}, \emptyset  , \emptyset )+ \\&w \cdot  (\epsilon_{\theta}(z_{t}, \mathcal{P}, I)-\epsilon _{\theta }(z_{t}, \emptyset  , \emptyset )),
		\end{split}
\end{equation}
where $z_{t}$ is the latent and $w$ is the interpolation coefficient which improves the alignment of generated samples with their conditioning. 
 The output $y_{c}$ of the ControlNet $\phi$ can be simplified as follows:

\begin{equation}
		\begin{split}
			\label{control_output}
                y_{c} = \sum_{j=1}^{J} \phi(z_{t},\mathcal{P},I; \theta_{b_{j}} ),
		\end{split}
	\end{equation}
where $J$ is the number of blocks in ControlNet and $\theta_{b_{j}}$ is the parameter of the $j-th$ block.


During the training of the ControlNet, visual and textual controls are perfectly alignment. However, When the prompt contains objects that are not spatial compositions of the visual control, the generated result may fail to depict these objects.
To address this issue, we initially generate object mask $\mathcal{OM}$, indicating the specific region in visual control that follows the textual prompt. Then we apply this mask to guide the ControlNet~\cite{zhang2023adding} to focus to this region, ensuring the extra prompt guidance such as a new object beyond the region is effective.

Concretely, we calculated the union of all object masks and obtained $\mathcal{OM}=(m_{1} \cup m_{2}...\cup m_{N})$. Then We recalculated $y_{c}$ with the below equation:

\begin{equation}
		\begin{split}
			\label{control_output_v1}
                y_{c} = \sum_{j=1}^{J} dm_{j} \cdot \phi_{2}(z_{t},\mathcal{P},I; \theta_{b_{j}} ),
		\end{split}
	\end{equation}
where $dm_{j}$ is the reshape form of $\mathcal{OM}$ and is adapted to the resolution of the $j-th$ block. 
Through this approach, ControlNet~\cite{zhang2023adding} selectively disregards features associated with misaligned visual elements outside of $\mathcal{OM}$. This exclusion facilitates precise spatial control by ControlNet~\cite{zhang2023adding} over specific image components and enables SD to generate objects as per the misaligned text descriptions. Our experimental results show that the application of masks to ControlNet features substantially mitigates conflicts between mismatched textual and visual controls, effectively addressing the problem of object missing in generated images

\subsection{Attribute-Matching Loss}


To improve attribute matching under visual controls, we employ various loss functions that iteratively denoise and update the noise map $z_{t}$ at each time step $t=(T,T-1,...,1)$ in the denoising process. We use spaCy’s transformer-based dependency parser to analyze the text prompt $\mathcal{P}$, extracting a set of attribute-object pairs $S=\sum_{i=1}^{N_{o}} (a_{i}, o_{i})$, where $N_{o}$ is the total number of objects, and $o_{i}$ and $a_{i}$ denote the word indices of the $i-th$ object and its corresponding attribute, respectively. The set $S$ is then divided into two subsets: $s_{1}$ which aligns with the visual control containing $N$ objects, and $s_{2}$, corresponding to the remaining $N_{o}-N$ objects in the misaligned portion of the prompt.

Our aim is to achieve semantic alignment between attribute words $a_{i}$ and their corresponding objects $o_{i}$, while maintaining the integrity of the image condition's control.
To this end, as shown in~\cref{overview}, we design two specific attribute-matching-loss functions: Language-guided loss and Mask-guided loss. These functions facilitate the alignment of $o_{i}$ and $a_{i}$ through cross-attention maps.
Specifically, during the forward process, we can obtain the cross-attention maps $A$ in SD and ControlNet named $A_{s}$ and $A_{c}$ respectively. For the $i-th$ prompt token $p_{i}$, the cross-attention map $A_{i}$ is calculated as follows:
\begin{equation}
		\begin{split}
			\label{cross-attention}
               A_{i}= softmax(\frac{QK_{i}^\top}{\sqrt{d}}),
		\end{split}
	\end{equation}
where the query $Q$ is derived from $z_{t}$, the key $K_{i}$ is derived from the word embedding of $p_{i}$, and $d$ is the latent dimension of of $Q$ and $K_{i}$.
$A_{s}$ and $A_{c}$ indicate the relevance between the given text prompt words and visual pixels. Moreover, $A_{c}$ extra indicates the relevance between the visual control and visual pixels.

Our main idea is to maximize the overlap of the attentive areas of the attribute-object pairs $(a_{i}, o_{i})$ in $s$, on the condition of $I$ and object masks $\mathcal{OM}$. 
To this end, we have formulated a loss function, designated as $L_{I}$. This function refines the attribute cross-attention maps in For pairs in $A_{s}$ and object cross-attention maps in $A_{c}$ for pairs in subset $s_{1}$. The goal is to progressively align these maps more closely, while distancing them from unrelated cross-attention maps, represented as $\bar{A_{c}}$. For the pairs in subset $s_{2}$, our approach primarily aligns their cross-attention maps within $A_{s}$ ensuring effective semantic alignment.

\begin{equation}
		\begin{split}
			\label{cross-attention}
                 &L_{I}=\sum_{(a_{i},o_{i}) \in s_{1}} dist(A_{c}(o_{i}), A_{s}(a_{i})) \\&-\sum_{ (a_{i},o_{i}) \in s_{1}} \sum_{w \in (a_{i},o_{i})} \sum_{A_{u}\in \bar{A_{c}}}  \frac{1}{|\bar{A_{c}}|} dist(A_{u}, A_{s}(w)) \\& + \sum_{(a_{i},o_{i}) \in s_{2}} dist(A_{s}(o_{i}), A_{s}(a_{i})),
		\end{split}
	\end{equation}
 \begin{equation}
		\begin{split}
			\label{cross-attention}
                dist(A_{i}, A_{j})= \frac{1}{2} D_{KL}(A_{i}||A_{j}) + \frac{1}{2} D_{KL}(A_{j}||A_{i}),
		\end{split}
	\end{equation}
where $dist$ is a measure function indicating the distance between the attention maps and $D_{KL}(A_{i}||A_{j})= {\textstyle \sum_{pixels}}A_{i}log(A_{i}/A_{j})$.

Furthermore, to meet the condition $\mathcal{M}$, another loss termed $L_{\mathcal{M}}$ is designed to constrain the target objects and attributes attentive area to the predefined object masks.
\begin{equation}
		\begin{split}
			\label{cross-attention}
                L_{\mathcal{M}}= - \sum_{i=1}^{N} \sum_{w\in s_{1,i}} \sum_{m\in \mathcal M_{i}} (A_{s}(w,m)-A_{s}(w,\bar{m} )),
		\end{split}
	\end{equation}
where $A_{s}(w, m)$ is a value of a cross-attention map in $A_{s}$ for word $w$ in $s_{1,i}$ at pixel $m$ in the $i-th$ object mask.


Then, all the losses are aggregated and back-propagation is computed to update the $z_{t}$ as follows:
\begin{equation}
		\begin{split}
			\label{cross-attention}
                z_{t}{'} \overset{}{\leftarrow} z_{t}- \alpha \nabla _{z_{t}}(l_{I},l_{\mathcal{M}}).
		\end{split}
\end{equation}

\begin{algorithm}[!t]    
	\caption{MGPF: Training-Free Prompt Following with Visual Controls}
	\textbf{Input:} A prompt $\mathcal{P}$, a visual control $I$, corresponding semantic masks $\mathcal{M}$, the source and target initial latent noise maps $z^s_T$ and $z_T$. \\
	\textbf{Output:} Latent map $z^s_0$, updated latent map $z_0$ corresponding to all the inputs.
	\begin{algorithmic}[1]
		\STATE $S= \{(a_{i}, o_{i})\}_{i=1}^{N} \leftarrow \text{Prompt Parser} (\mathcal{P})$
		\FOR{$t = T, T-1, ..., 1$}
		\STATE $\epsilon_s, A_{s}\leftarrow \epsilon_\theta(z^s_t, \mathcal{P},t), A_{c}\leftarrow \epsilon_{\phi}(z^s_t, \mathcal{P},I,t)$; 
		\STATE $z^s_{t-1} \leftarrow \text{Sample}(z^{s}_{t}, \epsilon_s)$;
		\STATE $ z_{t}{'} \hspace{-1mm}\leftarrow \text{MGPF}(A_{s},A_{c},\mathcal{M}, S)$;
		\STATE $\epsilon = \epsilon_{\theta,\phi}(z_{t}{'}, \mathcal{P},I,t)$;
		\STATE $z_{t-1} \leftarrow \text{Sample}(z_{t}, \epsilon)$;
		\ENDFOR
	\end{algorithmic}
	\label{alg:masactrl}
	\textbf{Return} $z^s_0, z_0$
\end{algorithm}

 Our algorithm can be summarized as~\cref{alg:masactrl}, which requires no training.

\section{Experiments}
\begin{table*}[ht!]
  \centering
  \caption{Quantitative results of all baseline models. We give results under four conditions (soft edge, canny edge, depth, segmentation). VQA-F, VQA-B, text-text, image-text represent LLM foreground VQA score, LLM background VQA score, clip text similarity between input text prompt and blip captions, clip image-text similarity respectively. Our method achieves the best results in all metrics.}
  \label{table:main}
  
  \resizebox{1\textwidth}{!}{
  \begin{tabular}{@{}l|ccccc|ccccc}
    \toprule
    Method & \multicolumn{5}{c|}{\textbf{Soft edge}} & \multicolumn{5}{c}{\textbf{Canny edge}} \\
     & VQA-AM & VQA-OG & Text-text & Image-text & Aesthestic & VQA-AM & VQA-OG & Text-text & Image-text & Aesthetic \\
    \midrule
    SD~\cite{rombach2022high} & 0.6321 \small \textcolor{red}{(-22.8\%)} & 0.4901 \small \textcolor{red}{(-27.0\%)} & 0.5816 \small \textcolor{red}{(-10.11\%)} & 0.2737 \small\textcolor{red}{(-9.4\%)} & \bf 5.24 \small\textcolor{blue}{(+0.9\%)} & 0.6561 \small\textcolor{red}{(-23.2\%)} & 0.4600 \small\textcolor{red}{(-30.9\%)}  & 0.5778 \small\textcolor{red}{(-11.5\%)} & 0.2738 \small\textcolor{red}{(-10.3\%)} & 5.20 \small\textcolor{blue}{(+0.3\%)}\\
    SD(mask)~\cite{rombach2022high} & 0.5596 \small\textcolor{red}{(-31.6\%)} & 0.4513 \small\textcolor{red}{(-32.7\%)} & 0.5872 \small\textcolor{red}{(-9.2\%)} & 0.2793 \small\textcolor{red}{(-7.5\%)} & \bf 5.24 \small\textcolor{blue}{(+0.9\%)} & 0.6423 \small\textcolor{red}{(-24.8\%)} & 0.4917 \small\textcolor{red}{(-26.2\%)} & 0.6151 \small\textcolor{red}{(-5.8\%)} & 0.2878 \small\textcolor{red}{(-5.7\%)} & \bf 5.31 \small\textcolor{blue}{(+2.5\%)} \\
    A\&E~\cite{chefer2023attend} & 0.7136 \small\textcolor{red}{(-12.8\%)} & 0.5316 \small\textcolor{red}{(-20.8\%)} & 0.5624 \small\textcolor{red}{(-13.1\%)} & 0.2717 \small\textcolor{red}{(-10.0\%)} & 5.08 \small\textcolor{red}{(-2.1\%)} & 0.7033 \small\textcolor{red}{(-17.6\%)} & 0.4903 \small\textcolor{red}{(-26.4\%)} & 0.5675 \small\textcolor{red}{(-13.1\%)} & 0.2716 \small\textcolor{red}{(-11.0\%)} & 5.02 \small\textcolor{red}{(-3.1\%)} \\
    BoxDiff~\cite{xie2023boxdiff} & 0.6426 \small\textcolor{red}{(-21.5\%)} & 0.5104 \small\textcolor{red}{(-23.9\%)} & 0.5655 \small\textcolor{red}{(-12.6\%)} & 0.2717 \small\textcolor{red}{(-10.0\%)} & 5.21 \small\textcolor{blue}{(+0.4\%)} & 0.6647 \small\textcolor{red}{(-22.1\%)} & 0.4841 \small\textcolor{red}{(-27.3\%)} & 0.5710 \small\textcolor{red}{(-12.6\%)} & 0.2741 \small\textcolor{red}{(-10.2\%)} & 5.15 \small\textcolor{red}{(-0.6\%)} \\
    Structure~\cite{feng2022training} & 0.6501 \small\textcolor{red}{(-20.6\%)} & 0.4899 \small\textcolor{red}{(-27.0\%)} & 0.5685 \small\textcolor{red}{(-12.1\%)} & 0.2733 \small\textcolor{red}{(-9.5\%)} & \bf 5.24 \small\textcolor{blue}{(+0.9\%)} & 0.6535 \small\textcolor{red}{(-23.5\%)} & 0.4594 \small\textcolor{red}{(-31.0\%)} & 0.5767 \small\textcolor{red}{(-11.7\%)} & 0.2730 \small\textcolor{red}{(-10.5\%)} & 5.20 \small\textcolor{blue}{(+0.4\%)} \\
    Linguistic~\cite{rassin2023linguistic} & 0.7556 \small\textcolor{red}{(-7.7\%)} & 0.5171 \small\textcolor{red}{(-23.0\%)} & 0.5854 \small\textcolor{red}{(-9.5\%)} & 0.2786 \small\textcolor{red}{(-7.8\%)} & 5.16 \small\textcolor{red}{(-0.6\%)} & 0.7482 \small\textcolor{red}{(-12.4\%)} & 0.4946 \small\textcolor{red}{(-25.7\%)} & 0.5869 \small\textcolor{red}{(-10.2\%)} & 0.2777 \small\textcolor{red}{(-9.0\%)} & 5.15 \small\textcolor{red}{(-0.6\%)} \\
    \midrule
    Ours & \textbf{0.8186} & \textbf{0.6710} & \textbf{0.6470} & \textbf{0.3020} & 5.19 & \textbf{0.8537} & \textbf{0.6658} & \textbf{0.6532} & \textbf{0.3051} & 5.18 \\
    \bottomrule
  \end{tabular}}

  \vspace*{0.05 cm} 

\resizebox{1\textwidth}{!}{
  \begin{tabular}{@{}l|ccccc|ccccc}
    \toprule
    Method & \multicolumn{5}{c|}{\textbf{Depth}} & \multicolumn{5}{c}{\textbf{Segmentation}} \\
     & VQA-AM & VQA-OG & Text-text & Image-text & Aesthetic & VQA-AM & VQA-OG & Text-text & Image-text & Aesthetic \\
    \midrule
    SD~\cite{rombach2022high} & 0.6638 \small\textcolor{red}{(-20.8\%)} & 0.5790 \small\textcolor{red}{(-11.7\%)} & 0.5963 \small\textcolor{red}{(-9.1\%)} & 0.2849 \small\textcolor{red}{(-6.7\%)} & 5.30 \small\textcolor{blue}{(+1.1\%)} & 0.6265 \small\textcolor{red}{(-20.3\%)} & 0.5613 \small\textcolor{red}{(-16.6\%)} & 0.6051 \small\textcolor{red}{(-7.8\%)} & 0.2870 \small\textcolor{red}{(-5.3\%)} & 5.36 \small\textcolor{blue}{(+0.9\%)} \\
    SD(mask)~\cite{rombach2022high} & 0.5953 \small\textcolor{red}{(-28.9\%)} & 0.4454 \small\textcolor{red}{(-32.1\%)} & 0.5992 \small\textcolor{red}{(-8.7\%)} & 0.2828 \small\textcolor{red}{(-7.4\%)} & 5.27 \small\textcolor{blue}{(+0.6\%)} & 0.5910 \small\textcolor{red}{(-24.8\%)} & 0.5159 \small\textcolor{red}{(-23.3\%)} & 0.6135 \small\textcolor{red}{(-6.6\%)} & 0.2896 \small\textcolor{red}{(-4.4\%)} & \bf 5.36 \small\textcolor{blue}{(+0.9\%)} \\
    A\&E~\cite{chefer2023attend} & 0.7150 \small\textcolor{red}{(-14.6\%)} & 0.5960 \small\textcolor{red}{(-9.2\%)} & 0.5775 \small\textcolor{red}{(-12.0\%)} & 0.2817 \small\textcolor{red}{(-7.8\%)} & 5.07 \small\textcolor{red}{(-3.2\%)} & 0.7195 \small\textcolor{red}{(-8.5\%)} & 0.5794 \small\textcolor{red}{(-13.9\%)} & 0.6001 \small\textcolor{red}{(-8.6\%)} & 0.2889 \small\textcolor{red}{(-4.7\%)} & 5.22 \small\textcolor{red}{(-1.7\%)} \\
    BoxDiff~\cite{xie2023boxdiff} & 0.6556 \small\textcolor{red}{(-21.7\%)} & 0.5986 \small\textcolor{red}{(-8.8\%)} & 0.5863 \small\textcolor{red}{(-10.7\%)} & 0.2831 \small\textcolor{red}{(-7.3\%)} & 5.24 \small\textcolor{blue}{(0.0\%)} & 0.6574 \small\textcolor{red}{(-16.4\%)} & 0.6007 \small\textcolor{red}{(-10.7\%)} & 0.6043 \small\textcolor{red}{(-8.0\%)} & 0.2879 \small\textcolor{red}{(-5.0\%)} & 5.31 \small\textcolor{blue}{(0.0\%)} \\
    Structure~\cite{feng2022training} & 0.6630 \small\textcolor{red}{(-20.9\%)} & 0.5784 \small\textcolor{red}{(-11.8\%)} & 0.5950 \small\textcolor{red}{(-9.3\%)} & 0.2850 \small\textcolor{red}{(-6.7\%)} & \bf 5.30 \small\textcolor{blue}{(1.1\%)} & 0.6254 \small\textcolor{red}{(-20.4\%)} & 0.5602 \small\textcolor{red}{(-16.7\%)} & 0.6053 \small\textcolor{red}{(-7.8\%)} & 0.2864 \small\textcolor{red}{(-5.5\%)} & 5.35 \small\textcolor{blue}{(0.8\%)} \\
    Linguistic~\cite{rassin2023linguistic} & 0.7477 \small\textcolor{red}{(-10.7\%)} & 0.6031 \small\textcolor{red}{(-8.1\%)} & 0.6042 \small\textcolor{red}{(-7.9\%)} & 0.2869 \small\textcolor{red}{(-6.1\%)} & 5.26 \small\textcolor{blue}{(0.4\%)} & 0.7734 \small\textcolor{red}{(-1.6\%)} & 0.6703 \small\textcolor{red}{(-0.4\%)} & 0.6493 \small\textcolor{red}{(-1.1\%)} & 0.3004 \small\textcolor{red}{(-0.9\%)} & 5.31 \small\textcolor{blue}{(0.0\%)} \\
    \midrule
    Ours & \textbf{0.8376} & \textbf{0.6560} & \textbf{0.6562} & \textbf{0.3055} & 5.24 & \textbf{0.7861} & \textbf{0.6728} & \textbf{0.6565} & \textbf{0.3030} & 5.31 \\
    \bottomrule
  \end{tabular}}
\end{table*}


In this section, we evaluate the effectiveness of our method in enhancing prompt following, focusing on object generation and attribute matching under visual controls. In~\cref{setup}, we delve into the specifics of our implementation details, the datasets, and evaluation metrics utilized in our experiments. Additionally, we demonstrate the efficiency of our method with both quantitative and qualitative results. Extensive experiments prove that our method can significantly boost the semantic alignment in text-to-image generation. 

\subsection{Evaluation Setup}
~\label{setup}

\noindent\textbf{Benchmarks.}
Considering the absence of established benchmarks for the evaluation of object missing and attribute mismatch under diverse visual controls, we construct a new benchmark. Our experiments involve four widely-used image conditions: depth, canny edge, soft edge, and segmentation. Following A\&E's approach~\cite{chefer2023attend, rassin2023linguistic}, we select images from the COCO dataset including 80 object types categorized into general, fruit, and animal groups. To compose our prompts, these objects are combined with 11 color-related attributes. The prompts are constructed with 1 or 2 objects in the image and randomly selected attributes. Additionally, to evaluate object generation capabilities beyond visual controls, we incorporated textual prompts from the COCO dataset into our prompts. Moreover, we consider two types of prompt: (1) a [color A] [object A], [text prompt beyond visual control], and (2) a [color A] [object A] and a [color B] [object B], [text prompt beyond visual control] and yield a total of 1384 samples. Details of different objects, attributes, and construction methods can be found in the supplementary material. 

\noindent\textbf{Evaluation Metrics.}
To quantitatively evaluate the performance of our method, we follow existing works~\cite{chefer2023attend} by utilizing text-text similarity and CLIP \cite{radford2021learning} image-text distances. Considering these two metrics are insufficient for evaluating object generation beyond visual control and attribute matching, we adopt the recently introduced VQA-based metric \cite{huang2023t2i}. This metric use task-related information as questions and evaluate the probability of a ``yes" response from our VQA model. For attribute matching, the question is designed as ``A [attribute] [object]?" for each pair in the text prompt, with the cumulative ``yes" probability serving as the metric. With respect to object generation beyond visual control, the question is ``[text prompt beyond visual condition]?", and its ``yes" probability is also computed. We refer to these metrics as VQA-AM (Attribute Matching) and VQA-OG (Object Generation), respectively. Additionally, based on concerns that some attention based methods might degrade image quality, we employ an aesthetic metric \cite{schuhmann2022laion} to evaluate the aesthetics of the generated images. We also conduct human evaluations for further validation.



\noindent\textbf{Baselines.} 
We compare our method with several training-free methods that intend to improve object generation and attributes matching. We re-implement these methods integrated with ControlNet~\cite{zhang2023adding}, including: Stable diffusion~\cite{rombach2022high}, A\&E~\cite{chefer2023attend}, Structure~\cite{feng2022training}, Lingustic~\cite{rassin2023linguistic}, BoxDiff~\cite{xie2023boxdiff} and Stable Diffusion\&mask~\cite{rombach2022high}. For BoxDiff, we calculate object bounding boxes from their masks, as an additional input. In Stable Diffusion\&mask, we utilize object masks to directly modify visual controls, integrating these with prompts for image generation.

\subsection{Results}
\begin{figure*}
  \centering
  \includegraphics[width=1\linewidth]{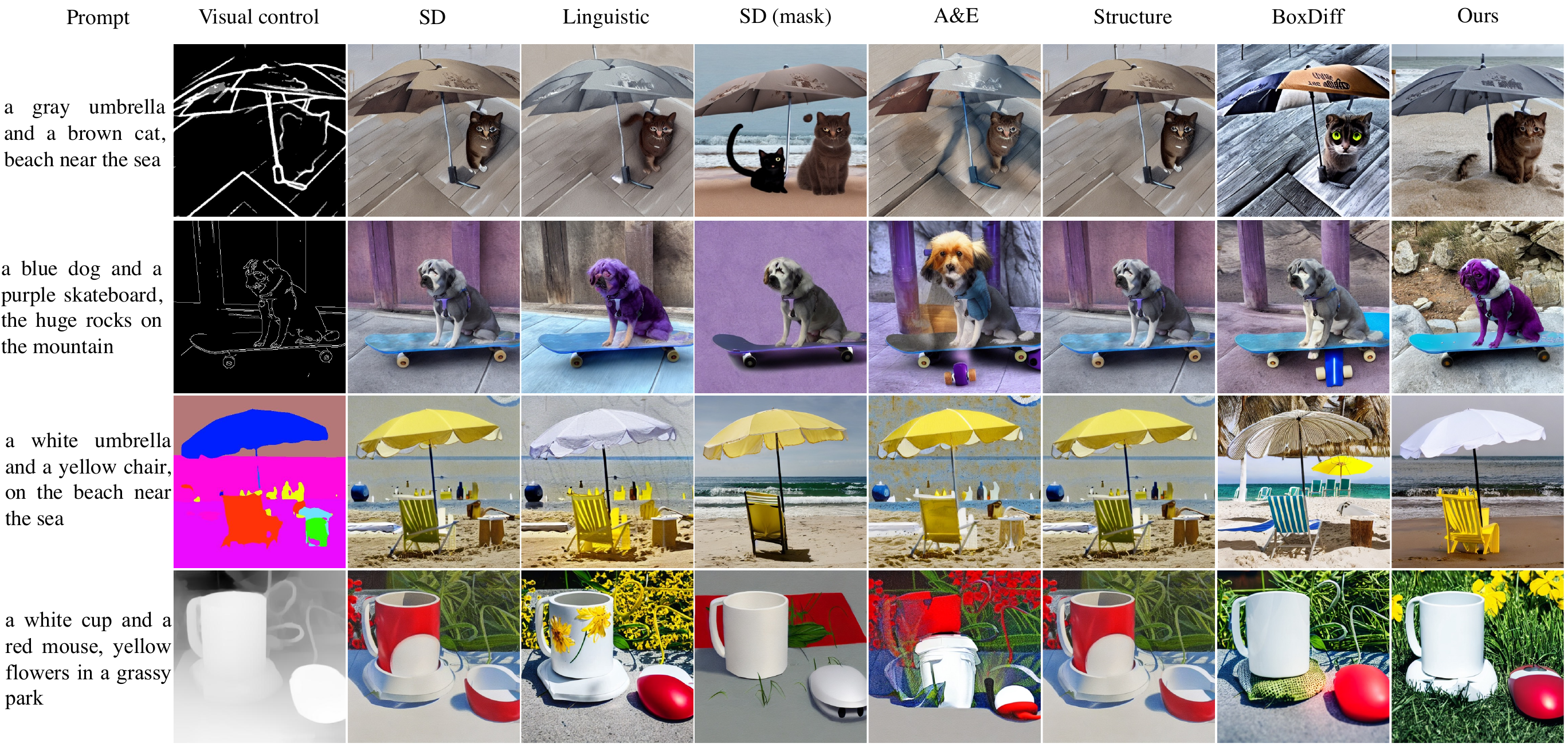}
  \caption{Qualitative comparison using prompts from our dataset. We show images generated by all our baseline methods. We use the same seed across all approaches.}
  \label{fig:showcase}
\end{figure*}

\noindent\textbf{Quantitative Results.}
We evaluated the effectiveness of every competing method using the aforementioned text-text, text-image, VQA-AM, VQA-OG, and aesthetic metrics. We demonstrate superior performance of our method across depth, soft edge, canny edge, and segmentation conditions, as detailed in \cref{table:main}. For the widely used text-text and image-text similarity metrics, our approach significantly outperforms all baseline models in all four conditions, demonstrating unparalleled prompt consistency. Considering that these two metrics do not entirely capture the ability of attribute matching and object generation beyond visual control, we further calculate the VQA-AM and VQA-OG metrics. For VQA-AM, our method remains superior to all baselines. Specifically, it outperforms the SD \cite{rombach2022high}+ControlNet and the most robust Linguistic \cite{rassin2023linguistic} baseline by 22.78\% and 7.7\% under the soft edge condition, respectively. (23.15\% and 12.36\% for canny edge; 20.75\% and 10.73\% for depth; 20.30\% and 1.62\% for segmentation). This demonstrates the superiority of our method to align attributes in the text prompt with corresponding objects in visual controlled scenarios. Regarding VQA-OG, which reflects object generation beyond visual control, our approach also achieved the best performance, surpassing the strongest baseline by 22.94\% under soft edge conditions (25.71\% for canny edge; 8.06\% for depth; 0.37\% for segmentation). Additionally, to address concerns that cross-attention based methods might affect image quality, we evaluate the aesthetic scores and find that all baseline methods exhibit comparable aesthetic scores and our method do not degrade the image quality. 

\begin{figure*}
  \centering
  \includegraphics[width=0.85\linewidth]{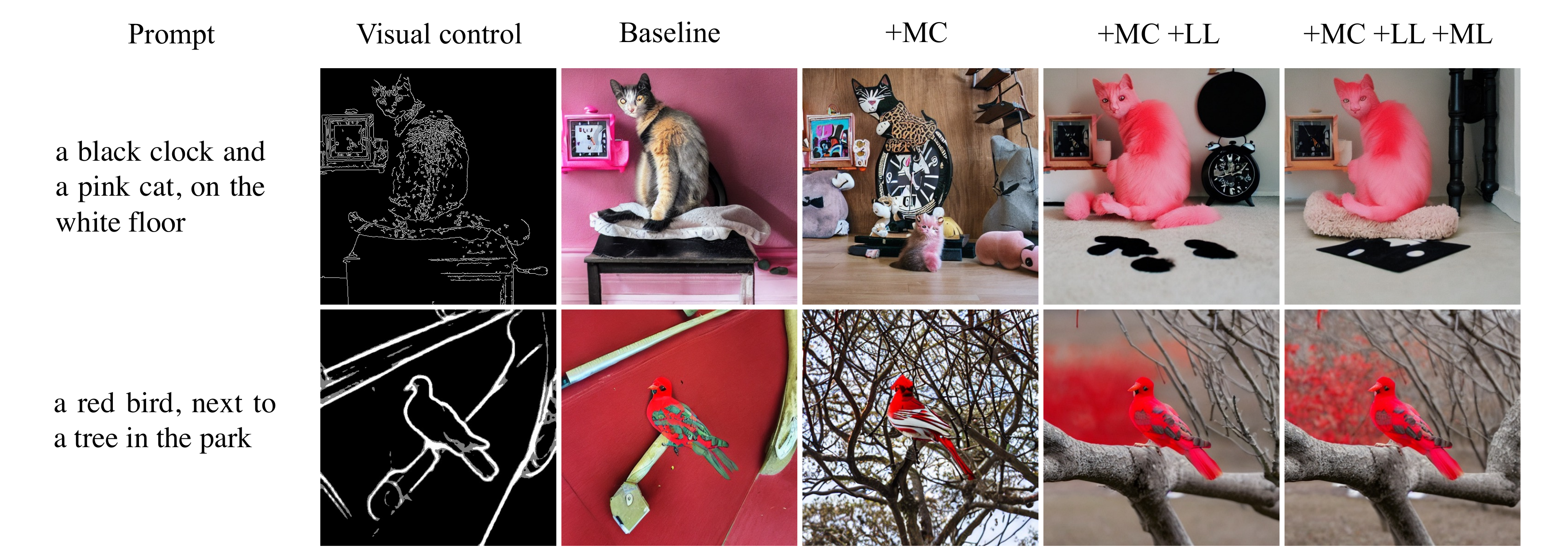}
  \caption{Qualitative ablation results. MC, ML, and LL denote Masked Controlnet, Mask Loss, and Language-Guided Loss respectively.}
  \label{fig:ablation}
\end{figure*}
\noindent\textbf{Qualitative Results.}
\cref{fig:showcase} illustrates a qualitative comparison of all baseline methods. Our analysis focuses on two aspects: the quality of attribute matching and the effectiveness of object generation beyond visual control. Regarding attribute matching, we can see that most baseline methods fail to achieve precise binding of attributes. While the Linguistic+ControlNet \cite{rassin2023linguistic} baseline occasionally facilitates attribute matching, it also produces artifacts, owing to the attention map shifting for both attributes and objects in the U-Net. In contrast, our method effectively binds attributes to their corresponding objects while preserving image quality, demonstrating the effectiveness of our attribute match losses. In terms of object generation beyond visual control, our method accurately produces objects in the misaligned portion of the prompt, outperforming other baselines. Notably, these models fail to maintain consistency with the text prompt, even when directly masking the misaligned part of the input visual control.

\begin{table}
  \centering
  \caption{Ablation studies of each module in our method, we conduct experiments under canny edge condition and report text-text, image-text, VQA-AM, VQA-OG metrics. ML: Mask-guided Loss, MC: Masked ControlNet, LL: Language-Guided Loss.}
  \label{table:ablations}
  \setlength{\tabcolsep}{3pt} 
  \resizebox{0.48\textwidth}{!}{
  \begin{tabular}{@{}l|cccc|}
  \toprule
  Method & \multicolumn{4}{c|}{\textbf{Canny edge}} \\ 
  & VQA-AM & VQA-OG & Text-text & Image-text \\
  \midrule
  Baseline & 0.6561 & 0.4600 & 0.5778 & 0.2738 \\
  +ML & 0.7715\textcolor{blue}{(+17.6\%)} & 0.4792 \textcolor{blue}{(+4.2\%)} & 0.5936 \textcolor{blue}{(+2.7\%)} & 0.2783 \textcolor{blue}{(+1.6\%)} \\
  +MC & 0.5845\textcolor{red}{(-10.9\%)} & 0.6139 \textcolor{blue}{(+33.5\%)} & 0.5817\textcolor{blue}{(+0.7\%)} & 0.2832\textcolor{blue}{(+3.4\%)} \\ 
  +LL & 0.8453\textcolor{blue}{(+28.8\%)} & 0.5218\textcolor{blue}{(+13.4\%)} & 0.6067\textcolor{blue}{(+5.0\%)} & 0.2822\textcolor{blue}{(+3.1\%)} \\ 
  +ML+LL & \textbf{0.8554}\textcolor{blue}{(+30.4\%)} & 0.5175\textcolor{blue}{(+12.5\%)} & 0.6080\textcolor{blue}{(+5.2\%)} & 0.2828\textcolor{blue}{(+3.3\%)} \\
  +ML+MC & 0.7791\textcolor{blue}{(+18.7\%)} & 0.6529\textcolor{blue}{(+42.0\%)} & 0.6367\textcolor{blue}{(+10.2\%)} & 0.3017\textcolor{blue}{(+10.2\%)} \\
  +MC+LL & 0.8492\textcolor{blue}{(+29.4\%)} & \textbf{0.6738}\textcolor{blue}{(+46.5\%)} & 0.6524\textcolor{blue}{(+12.9\%)} & 0.3040\textcolor{blue}{(+11.0\%)} \\
  +MC+LL+ML & 0.8537\textcolor{blue}{(+30.1\%)} & 0.6658\textcolor{blue}{(+44.7\%)} & \textbf{0.6532}\textcolor{blue}{(+13.1\%)} & \textbf{0.3051}\textcolor{blue}{(+11.4\%)} \\
  \bottomrule
  \end{tabular}}
\end{table}

\noindent\textbf{Ablation Study.} Initially, we quantify the impact of each component within our method as shown in ~\cref{table:ablations}. We perform experiments under the canny edge condition and use SD~\cite{rombach2022high}+ControlNet as the baseline. The incorporation of Language-guided Loss (LL) and Mask-guided Loss (ML) improves the baseline model across all metrics, particularly in VQA-AM, proving the effectiveness of these losses in enhancing attribute matching. Notably, Masked ControlNet (MC) improves the VQA-OG by a large margin(33.5\%), but affected the VQA-AM (decreasing by 10.9\%). We observed that combining LL and ML achieves the best results in VQA-AM, indicating superior attribute matching. Integrating LL with MC, we obtain the highest VQA-OG score. Finally, with all three components, our method achieves the balance results in both VQA-AM and VQA-OG metrics. Additionally, ~\cref{fig:ablation} shows the result of progressively integrating our modules into the base SD \cite{rombach2022high}+ControlNet model. 
The addition of MC improves object generation in the latter part of the text prompt. For instance, in the second row, the image distinctly features a tree in the park.
Integrating LL enhances attribute matching (e.g., a pink cat and black clock in the first row, and a red bird in the second), though it introduced some artifacts in the generated images, such as an additional block in the first case and red color artifacts in the second.
When further combining ML, we solve the artifacts, achieving the best performance in both attribute matching and object generation.





\begin{figure}
  \centering
  \includegraphics[width=0.98\linewidth]{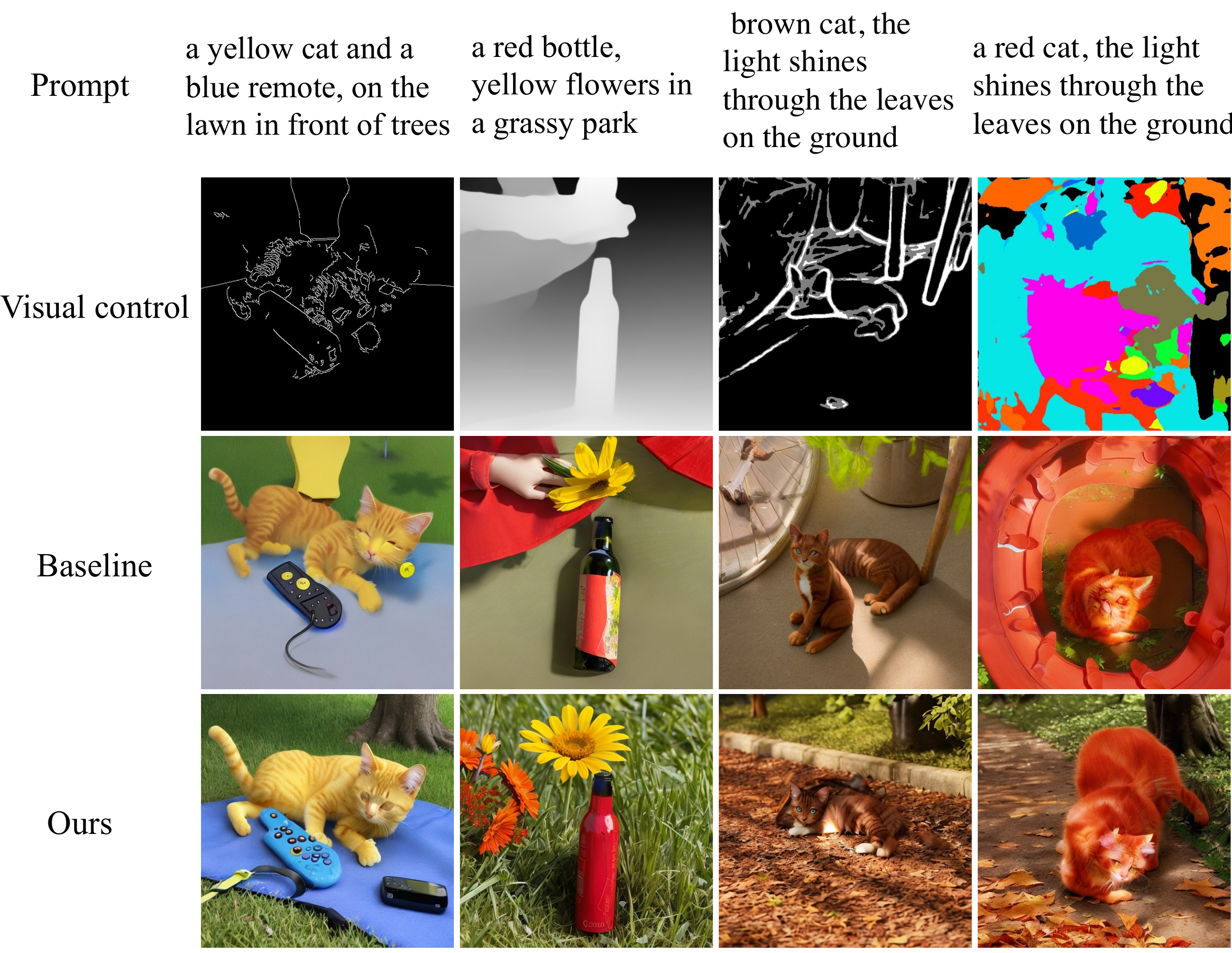}
  \caption{Qualitative Comparison using ChilloutMix, we compare our method with baseline model.}
  \label{fig:chillout}
  \vspace{-0.2cm}
\end{figure}

\subsection{Robustness to Other Models: ChilloutMix}

Additionally, we have cooperated our method to other popular models, specifically ChilloutMix\footnote{\url{https://civitai.com/models/6424}}. \cref{fig:chillout} shows the visualization results of our approach and the SD \cite{rombach2022high}+ControlNet models with a fixed random seed. Our method effectively generates objects that extend beyond visual control and matches attributes to their corresponding objects. For example, in the first column, our approach accurately binds the attribute ``blue" to the ``remote", while the baseline method fails. Moreover, our method consistently produces better scenarios that more closely to prompts such as ``on the lawn in front of trees," thereby demonstrating the generalizability of our proposed approach. For more cases, please refer to the supplementary material.

\subsection{Human Evaluations}
\begin{table}
    \caption{Human Evaluation results compared with all baseline methods. AM, OG, VC and aesthetic represents attributes matching, object generation beyond visual control, visual control following and aesthetic respectively.}
   \label{tables:human}
  \centering
  \begin{tabular}{l|ccll}
    \toprule
    Model   & AM & OG & VC & aesthetic  \\
    \midrule
    Ours   &  129  & 67 & 170 & 118  \\
    Linguistic~\cite{rassin2023linguistic}  &  97 & 51 & 168 & 98 \\
    A\&E~\cite{chefer2023attend}  &  76 & 35 & 156 & 60 \\
    Structure~\cite{feng2022training}  &  71 & 45 & 171 & 116 \\
    boxdiff~\cite{xie2023boxdiff} &  82 & 38 & 161 & 88 \\
    SD~\cite{rombach2022high} &  72 & 42 & 172 & 111 \\
    SD(mask)~\cite{rombach2022high} & 74 & 54 & 172 & 109 \\
    No winner & 38 & 102 & 2 & 17 \\
    \bottomrule
  \end{tabular}
\end{table}
\vspace{-0.2cm}

To further validate our approach, we conduct a user study with images generated by our method and compare them to those from various baseline methods. survey respondents are asked to select the image set that best aligned with the input prompt under visual controls, focusing on Attribute Matching (AM), Object Generation Beyond Control (OG), Visual Controls Following (VC), and Aesthetics. We randomly sample 200 prompts from our constructed dataset, covering all four kinds of visual controls, and generate images using all methods. The results, presented in \cref{tables:human}, indicate the number of selections for each method. For AM and OG, the majority of survey respondents prefer the results produced by our method, thereby providing additional evidence for its effectiveness. Regarding VC and Aesthetics, we can see that our method neither disrupts the control of image conditions nor degrades the aesthetic quality of the generated images. Further details are available in the supplementary material.

\section{Conclusions}

In this work, we propose a training-free method for the prompt following under visual control. To enhance object generation, we introduce object masks and propose Masked ControlNet to handle the misalignment between the prompt and visual control. To tackle the issue of attribute mismatch, we propose two straightforward losses to align the cross-attentions between attribute and object pairs, achieving effective attribute binding. Experimental results demonstrate the superiority of our approach. Additionally, the integration of our method with any Diffusion-based model expands its range of applications and customization possibilities. 

One limitation of our method is that in the more complex scenarios, such as attribute match involving multiple or smaller objects under visual control, it is challenging to accurately bind attributes to these objects. This may arise from the low resolution of the current attention maps, limiting our capacity to perform precise localized attribute binding. To alleviate this, we suggest incorporating cross-attention also in higher-resolution layers. We will leave this for future works.

{\small
\bibliographystyle{ieee_fullname}
\bibliography{egbib}
}

\end{document}